# PyGALAX: An Open-Source Python Toolkit for Advanced Explainable Geospatial Machine Learning

Pingping Wang[1], Yihong Yuan[1*], Lingcheng Li[2], Yongmei Lu[1]

1 Department of Geography and Environmental Studies, Texas State University, San Marcos, TX, USA 2 Atmospheric, Climate, and Earth Sciences Division, Pacific Northwest National Laboratory, Richland, WA, USA

[*]Corresponding author: Yihong Yuan

GitHub Repository: https://github.com/Pingping9/PyGALAX

**Summary**

PyGALAX is a Python package for geospatial analysis that integrates automated machine learning (AutoML) and explainable artificial intelligence (XAI) techniques to analyze spatial heterogeneity in both regression and classification tasks. It automatically selects and optimizes machine learning models for different geographic locations and contexts while maintaining interpretability through SHAP (SHapley Additive exPlanations) analysis. PyGALAX builds upon and improves the GALAX framework (Geospatial Analysis Leveraging AutoML and eXplainable AI) (Wang, Yuan, Li, et al., 2025), which has proven to outperform traditional geographically weighted regression (GWR) methods. Critical enhancements in PyGALAX from the original GALAX framework include automatic bandwidth selection and flexible kernel function selection, providing greater flexibility and robustness for spatial modeling across diverse datasets and research questions.

PyGALAX not only inherits all the functionalities of the original GALAX framework but also packages them into an accessible, reproducible, and easily deployable Python toolkit while providing additional options for spatial modeling. It effectively addresses spatial non-stationarity and generates transparent insights into complex spatial relationships at both global and local scales, making advanced geospatial machine learning methods accessible to researchers and practitioners in geography, urban planning, environmental science, and related fields.

**Statement of need**

Understanding spatially heterogeneous and non-linear relationships is a persistent challenge in geography and environmental sciences, such as urban analytics, human mobility studies, and public health (Siddique, 2024; Wang & Yuan, 2025; Wang, Yuan, & Myint, 2025; Yuan, Wang, McKenzie, et al., 2025). Traditional approaches such as GWR provide valuable insights into spatial non-stationarity but are constrained by their linear assumptions and limited ability to handle complex, high-dimensional data (Brunsdon et al., 1996; Fotheringham et al., 2017). In contrast, modern machine learning techniques can model complex, non-linear relationships but often ignore spatial context and operate as "black boxes," leading to reduced interpretability and limited utilization in decision-making applications (Nagarajah & Poravi, 2019; Wang et al., 2021). While recent advances have introduced machine learning into spatial analysis, existing tools lack the flexibility needed for diverse research applications. For example, PyGRF (Sun et al., 2024) is a comparable Python package that implements the established Geographical Random Forests method for spatial analysis (Yuan, Wang, & Summers, 2025). While PyGRF advances spatial modeling by incorporating Random Forests into a geographic framework, it is constrained to a single algorithm and primarily focuses on regression tasks.

To address this methodological gap, GALAX was recently developed as a hybrid analytical framework that integrates AutoML and XAI within a spatial modeling structure (Wang, Yuan, Li, et al., 2025). GALAX enables researchers to automatically identify optimal machine learning algorithms for each geographic location and context (Wang et al., 2021), incorporate spatial weighting into the workflow (Brunsdon et al., 1996), and interpret both global and local relationships through SHAP-based explainability (Lundberg & Lee, 2017). This integration enables a transparent, adaptive, and data-driven understanding of spatially varying processes, representing a significant advancement beyond both traditional GWR and non-spatial machine learning approaches.

Building upon the success of the GALAX conceptual framework, PyGALAX is created to operationalize GALAX as an open-source Python package, while providing additional spatial analysis flexibility for practitioners and researchers. Specifically, PyGALAX enhances the original GALAX model through automatic bandwidth and kernel selection and a user-friendly implementation. These improvements make spatial AutoML workflows accessible to a wider research community and ensure reproducibility across studies. Furthermore, PyGALAX supports both regression and classification tasks, expanding its applicability to diverse application domains such as mobility analysis, environmental monitoring, and public health research, areas where understanding both the "what" and the "why" of spatial patterns is critical for evidence-based decision-making (Georganos et al., 2021; Li, 2022).

**Installation**

PyGALAX is compatible with Python 3.9 and later and can be easily installed directly from its GitHub repository:

git clone https://github.com/Pingping9/PyGALAX

cd PyGALAX

pip install .

**Key features**

PyGALAX offers several distinctive capabilities that make it suitable for advanced spatial analysis (Figure 1 and Table 1):

- **Spatial AutoML integration:** PyGALAX automates the GALAX framework that implements geographically weighted AutoML, where different machine learning algorithms (e.g., Random Forest, XGBoost, Extra Trees) are automatically selected and optimized for each spatial location based on local data characteristics. This approach enables the capture of varying relationship structures across geographic space.
- **Fixed or Adaptive bandwidth selection:** PyGALAX provides multiple

bandwidth selection methods, including Incremental Spatial Autocorrelation (ISA) analysis (Sun et al., 2024) and performance-based optimization (Brunsdon et al., 1996), ensuring optimal spatial scale selection for different datasets and research objectives.
- **Explainable spatial AI:** Through SHAP integration, PyGALAX provides detailed explanations of model predictions, revealing how different features contribute to outcomes across geographic space. This includes both the importance of local features and the spatial patterns of variable influence.
- **Unified regression and classification:** Unlike many spatial analysis tools that focus exclusively on continuous outcomes, PyGALAX seamlessly handles both regression and classification tasks, making it versatile for diverse research applications.
- **Flexible kernel functions:** PyGALAX supports multiple spatial weighting schemes (e.g., bisquare, gaussian, exponential) and both fixed and adaptive bandwidth approaches (Brunsdon et al., 1996), allowing customization for different spatial processes.
- **Parallel processing:** Built-in support for multicore processing enables efficient analysis of large spatial datasets.

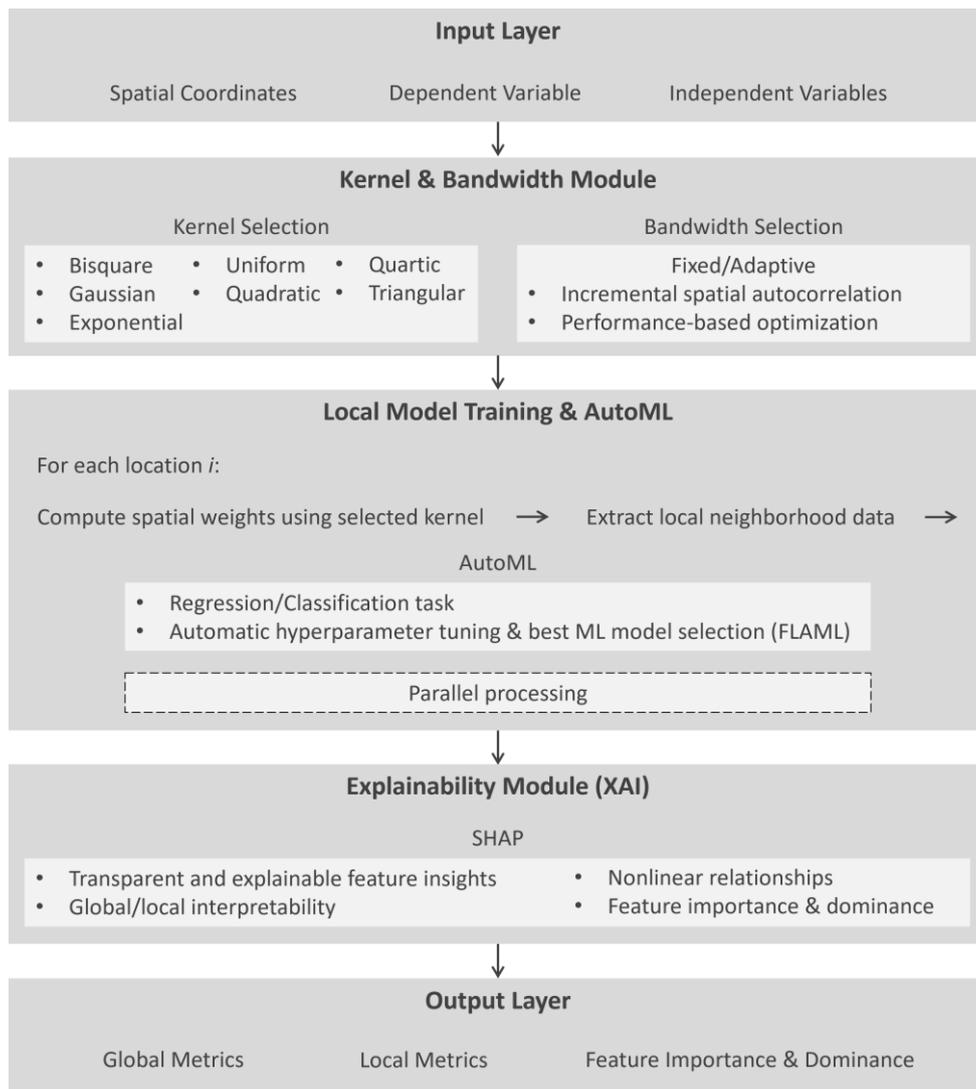

**Figure 1.** PyGALAX Methodological Framework Extended from Wang, Yuan, Li, et al. (2025)

**Table 1.** Example PyGALAX Commands

| Command | Description |
| --- | --- |
| Kernel(coords[i], coords, bw, fixed=False, function='bisquare') | Create a kernel weighting matrix for spatial dependence modeling. |
| search_bw_lw_ISA(X, y, coords) | Run standalone ISA to estimate an optimal bandwidth. |
| search_bandwidth(X, y, coords, automl_settings) | Perform bandwidth optimization using AutoML model performance metrics. |
| model = GALAX(coords, y, X, task='regression', bw=None, kernel='bisquare') | Initialize a GALAX model with coordinates (coords), target variable (y), and features (X). The model automatically selects the optimal bandwidth when bw=None. |
| results = model.fit() | Fit the model using either ISA-based or performance-based bandwidth optimization, depending on data and task type. |
| results.summary() | Display summary statistics for global and local model performance (e.g., $R^2$, RMSE, or accuracy, precision, recall, and F1 score). |
| results.save_results("pygalax_output.joblib") | Save the full model outputs (including SHAP values and predictions) as a serialized Joblib file. |
| results.get_detailed_shap_for_location(5) | Retrieve detailed SHAP interpretation for a specific spatial location. |

**Model architecture**

PyGALAX is built on established Python libraries, including scikit-learn, FLAML for AutoML functionality, and SHAP for explainability. The modular architecture allows for easy extension and customization while maintaining computational efficiency through joblib-based parallel processing.

PyGALAX follows object-oriented design principles with clear separation between spatial weighting (Kernel class), model fitting (GALAX class), and results analysis (GALAXResults class). This design facilitates both ease of use for standard applications and extensibility for advanced research needs.

PyGALAX's modular structure enables future enhancements in multiple directions. The architecture allows for integration of additional AutoML backends (such as AutoGluon) and alternative XAI beyond SHAP. In addition, it provides a foundation for extending PyGALAX handles spatiotemporal data through geographically and temporally weighted frameworks, and for incorporating multi-scale analysis capabilities that account for varying spatial scales of different geographic processes.

**Applications**

PyGALAX enables advanced spatial analysis across diverse research domains:

- **Urban and transportation planning:** Analyzing spatial patterns in human mobility (Wang, Yuan, Li, et al., 2025), identifying factors influencing travel behavior across different urban contexts, and understanding how infrastructure and socioeconomic variables affect transportation choices.
- **Environmental science and ecology:** Modeling spatial variations in environmental quality, examining relationships between land use patterns and

- ecosystem services, and predicting pollution distribution with location-specific drivers.
- **Public health:** Investigating geographic disparities in health outcomes, identifying spatially varying risk factors for disease transmission, and optimizing healthcare resource allocation based on local population characteristics.
- **Crime analysis:** Examining spatial inequality in crime patterns, analyzing neighborhood effects on criminal activity, and understanding how contextual factors influence crime rates across different regions.
- **Real estate and economics:** Predicting property values with spatially varying determinants, analyzing local market dynamics, and identifying factors driving economic development in different geographic areas.
- **Agriculture and natural resources:** Modeling crop yields with location-specific environmental and management factors, optimizing resource allocation based on spatial heterogeneity, and predicting soil properties across agricultural landscapes.
- **Tourism and recreation analytics:** Understanding spatial patterns in visitor flows, modeling heterogeneous factors influencing site attractiveness, and optimizing infrastructure placement in parks or cultural destinations.
- **Disaster resilience and emergency management:** Identifying spatially varying drivers of vulnerability, modeling community-level resilience, and optimizing evacuation or resource-deployment strategies based on local conditions.

## Acknowledgements

The first and second authors received funding support from the 2024-2025 Texas State University College of Liberal Arts Seed Grant.

# References

Brunsdon, C., Fotheringham, A. S., & Charlton, M. E. (1996). Geographically weighted regression: a method for exploring spatial nonstationarity. *Geographical analysis*, *28*(4), 281-298.

Fotheringham, A. S., Yang, W., & Kang, W. (2017). Multiscale geographically weighted regression (MGWR). *Annals of the American Association of Geographers*, *107*(6), 1247-1265.

Georganos, S., Grippa, T., Niang Gadiaga, A., Linard, C., Lennert, M., Vanhuysse, S., Mboga, N., Wolff, E., & Kalogirou, S. (2021). Geographical random forests: a spatial extension of the random forest algorithm to address spatial heterogeneity in remote sensing and population modelling. *Geocarto International*, *36*(2), 121-136.

Li, Z. (2022). Extracting spatial effects from machine learning model using local interpretation method: An example of SHAP and XGBoost. *Computers, Environment and Urban Systems*, *96*, 101845.

Lundberg, S. M., & Lee, S.-I. (2017). A unified approach to interpreting model predictions. *Advances in neural information processing systems*, *30*.

Nagarajah, T., & Poravi, G. (2019). A review on automated machine learning (AutoML) systems. 2019 IEEE 5th International Conference for Convergence in Technology (I2CT),

Siddique, I. (2024). Machine learning empowered geographic information systems: Advancing Spatial analysis and decision making. *World Journal of Advanced Research and Reviews*, *22*(1), 10.30574.

Sun, K., Zhou, R. Z., Kim, J., & Hu, Y. (2024). PyGRF: An improved Python Geographical Random Forest model and case studies in public health and natural disasters. *Transactions in GIS*, *28*(7), 2476-2491.

Wang, C., Wu, Q., Weimer, M., & Zhu, E. (2021). Flaml: A fast and lightweight automl library. *Proceedings of Machine Learning and Systems*, *3*, 434-447.

Wang, P., & Yuan, Y. (2025). Analyzing Key Factors Influencing Human Mobility Before and During COVID-19 With Explainable Machine Learning. *Transactions in GIS*, *29*(1), e13271.

Wang, P., Yuan, Y., Li, L., & Lu, Y. (2025). GALAX: A Framework for Geospatial Analysis Leveraging AutoML and eXplainable AI. *Annals of the American Association of Geographers*, 1–27. https://doi.org/10.1080/24694452.2025.2591684

Wang, P., Yuan, Y., & Myint, S. W. (2025). Developing mobility zones based on heat exposure and vegetation condition: a tale of two transportation modes. *International Journal of Digital Earth*, *18*(2), 2552878.

Yuan, Y., Wang, P., McKenzie, G., Sun, M., Xu, Y., & Verma, P. (2025). Human Mobility in the Postpandemic World. In *Urban Human Mobility: Practices, Analytics, and Strategies for Smart Cities* (Vol. 19, pp. 221-234). CRC Press. https://doi.org/10.1201/9781003503262-23

Yuan, Y., Wang, P., & Summers, L. (2025). Comparing the Spatiotemporal Patterns of Crime Incidents based on Time Series Analysis: A Case Study of Austin, Texas. Proceedings of the ICA,